\pdfoutput=1
\documentclass[11pt]{article}

\usepackage{EACL2023}

\usepackage{times}
\usepackage{latexsym}
\usepackage{booktabs}
\usepackage{lipsum}
\usepackage{graphicx}
\usepackage{amssymb}

\usepackage{tablefootnote}

\usepackage[T1]{fontenc}
\usepackage[utf8]{inputenc}

\usepackage{microtype}

\usepackage{inconsolata}

\title{\textsc{TextWorldExpress}: Simulating Text Games at \\One Million Steps Per Second}
\author{Peter Jansen \\
  University of Arizona, Tucson, USA \\
  \texttt{pajansen@arizona.edu} \\\And
  Marc-Alexandre Côté \\
  Microsoft Research Montréal \\
  \texttt{macote@microsoft.com} \\}

\begin{document}
\maketitle
\begin{abstract}
Text-based games offer a challenging test bed to evaluate virtual agents at language understanding, multi-step problem-solving, and common-sense reasoning. % in situated environments that are rendered entirely in text, as natural language descriptions. % Agents typically must learn to perform complex multi-step tasks that require both domain-specific and common-sense reasoning like navigation or instruction following.
However, speed is a major limitation of current text-based games, capping at 300 steps per second, mainly due to the use of legacy tooling.  In this work we present \textsc{TextWorldExpress}, a high-performance simulator that includes implementations of three common text game benchmarks that increases simulation throughput by approximately \textbf{three orders of magnitude}, reaching over one million steps per second on common desktop hardware.  This significantly reduces experiment runtime, enabling billion-step-scale experiments in about one day.\footnote{Code: \href{https://github.com/cognitiveailab/TextWorldExpress}{github.com/cognitiveailab/TextWorldExpress}} \footnote{Video: \href{https://youtu.be/HLFAnRKuTlE}{youtu.be/HLFAnRKuTlE}}
\footnote{Demo: \href{https://marccote-textworldexpress.hf.space}{marccote-textworldexpress.hf.space}}

\end{abstract}

\section{Introduction}

One of the long standing goals of artificial intelligence is to create agents that can work and reason in embodied environments.  Toward this goal, a variety of virtual environments have been created that allow simulated robots the opportunity to learn to a variety of tasks, in settings from household environments \cite{kolve2017ai2,ALFRED20} to Minecraft \cite{guss2019minerl}.  Because high-fidelity 3D virtual environments are challenging and resource intensive to develop, simpler 2D environments have also been proposed \citep[e.g.][]{chevalier2019babyai,kuttler2020nethack} that allow agents to focus on learning skills such as search or navigation in graphically simpler environments. 

Recently, text games -- or environments rendered entirely in natural language -- have emerged as an alternate research methodology for embodied agent research, centrally due to their low barrier to entry compared to 3D games, coupled with their ability to easily model complex tasks at a high-level \citep[see][for review]{jansen2022systematic}.  For example, a cooking game might require an agent to read a recipe, find ingredients, then prepare those ingredients to create a meal.  Text games model an agent as they navigate an environment, rendering their observations in text (e.g. \textit{``You are in the kitchen.  You see...''}).  Similarly, agents interact with the environment through abstracted high-level natural language commands (e.g. \textit{``move south''}, or \textit{``pick up carrot''}), rather than lower-level actions common in 3D environments (e.g. \textit{rotate agent 2 degrees clockwise}). % that change the state of the environment towards the goal of completing a task.  

%
% Table of speed comparisons
%
\begin{table}[t!]
    \centering
    \footnotesize
    \begin{tabular}{lcc}
    \toprule
         \textbf{Environment Simulator} & \textbf{SPS} \\
         \midrule
         \textit{2D/3D Simulators}\tablefootnote{Performance reported from \cite{zholus-iglu2022}.} \\
         \midrule
         ~~\textsc{AI2Thor}    \cite{kolve2017ai2}                       &   30$\dagger$     \\
         ~~\textsc{MineRL}     \cite{guss2019minerl}                     &   180$\dagger$     \\
         ~~\textsc{BabyAI}     \cite{chevalier2019babyai}                &   3k     \\
         ~~\textsc{NetHack}    \cite{kuttler2020nethack}                 &   14k   \\
         ~~\textsc{MegaVerse}  \cite{petrenko2021megaverse}              &   327k$\dagger$     \\    
         
         \midrule
         \textit{Text Game Simulators}\tablefootnote{Benchmark scripts provided in the code repository.} \\
         \midrule
         ~~\textsc{TextWorld}     \cite{Ct2018TextWorldAL}             &   300     \\
         ~~\textsc{Jericho}        \cite{Hausknecht2020InteractiveFG}  &   1       \\
         ~~\textsc{ScienceWorld}   \cite{Wang2022ScienceWorldIY}       &   20   \\
         \cmidrule(lr){1-2}
         ~~\textsc{TextWorldExpress} \textit{(online,  \textsc{Python})}                      &   32k    \\ 
         ~~\textsc{TextWorldExpress} \textit{(precrawled, \textsc{Python})}                   &   316k   \\  % 220k-316k (python) \\
         ~~\textsc{TextWorldExpress} \textit{(online, \textsc{JAVA})}                   &   212k   \\  % 155k-256k (scala) \\
         ~~\textsc{TextWorldExpress} \textit{(precrawled, \textsc{JAVA})}               &   4M   \\  % 220k-316k (python) \\
         \bottomrule
    \end{tabular}
    \caption{Single-thread simulation speed of common 2D, 3D, and text-game environment simulators.  Speed is measured in terms of Steps Per Second (SPS). $\dagger$ symbolizes that simulation is carried out on GPUs. \textsc{TextWorldExpress} outperforms other text game simulators by approximately three orders of magnitude.}
    \label{tab:speed-comparison}
\end{table}

Text games require a variety of common-sense knowledge to complete successfully~\cite{ryu-etal-2022-fire,murugesan-etal-2021-efficient}, including understanding common procedures (such as how to read and follow instructions), as well as affordances about the world -- for example, that buildings have rooms, containers must be opened before their contents can be observed or removed, and so forth.  As such, text games are still extremely challenging for agents, with current state-of-the-art performance at only 12\% for classic interactive fiction games such as \textit{Zork} \cite{yao-etal-2021-reading,ammanabrolu2021motivate}.  Similarly, interactivity and explicit step-by-step reasoning appears challenging for agents.   For example, there appears to be a large dissociation between a model's ability to answer questions about topics (e.g., science exam questions) and its ability to perform very similar experiments in interactive text environments, even with substantial training \cite{Wang2022ScienceWorldIY}.  This suggests that explicit interactive multi-step reasoning is still very challenging for contemporary methods like language models, and that accurate procedural knowledge is currently difficult to generate.
%For example, a model that can answer 90\% of multiple choice elementary science exam questions correctly fails to correctly solve text games that test that same knowledge, even with significant training \cite{Wang2022ScienceWorldIY}.  
Together, these highlight the importance of using text games as a vehicle for explicit, embodied, step-by-step reasoning about the world. 

To help support these efforts, a number of simulators have recently been developed for text game research, shown in Table~\ref{tab:speed-comparison}.  Current tooling for text games is built on legacy code bases, providing strong limitations in rendering speed -- at present, most simulators are limited to running at between 1 and 300 steps per second.  This generally limits agents from using modeling paradigms with fast iteration cycles and high sample requirements (such as reinforcement learning, or evolutionary learning), and restricts users to modeling techniques with large train and inference cycles (such as language models) where the simulator no longer becomes the bottleneck in experiment runtimes.

In this work, we develop a high-speed framework for text-based games in natural language processing research.  Our contributions are: 
\begin{enumerate}
    \item \textsc{TextWorldExpress}, a highly optimized simulator that includes reimplementations of three text game benchmarks focusing on instruction following, commonsense reasoning, and object identification, as well as other newer benchmarks for evaluating arithmetic, navigation, and neurosymbolic reasoning.
    \item We empirically demonstrate that this simulator runs up to \textit{three orders of magnitude} faster than current tooling, reaching 300k steps per second (SPS) on a single-thread, and exceeding 1M SPS on modest multi-core desktop hardware.  This substantially reduces experiment times (from weeks to hours) for sample-heavy machine learning agents.
\end{enumerate}
%% TODO (PJ): Sell as a more general framework, with implementations of N common benchmarks as well as other newer games?  Also mention ease of use in developing new games?

%- Mention valid action handicap -- that in spite of it being substantial, because text games are hard, nearly every agent uses them and still achieves very modest performance, with top performance for benchmarks such as Zork only at 0.12 (with this handicap). 
%- Throughput: At >125k steps/sec, allows 1M steps/sec on 8-core machine. 
%\todo{- TODO: Mention two speed modes: Normal/online-generation, or Precrawled/offline generation (export to JSON, speed as fast as a pointer lookup in memory). }
%\todo{- TODO: Implement Scala -> Python API, measure speed in Python vs Scala. }

\section{Related Work}
{\flushleft\textbf{Research Paradigm:}} Text games are a rapidly expanding research paradigm for learning and evaluating situated natural language processing agents on a variety of tasks, with over 100 papers written using this paradigm in the last few years \citep[see][for review]{jansen2022systematic}.  This may be in part due to language providing useful abstractions for more efficient exploration and planning \cite{Karch2020LanguageGoalIT, Colas2020LanguageAA,Mu2022ImprovingIE,Tam2022SemanticEF}, making task modeling at the level of language more easily approached than with lower-level 3D simulations.

{\flushleft\textbf{Agent Modeling:}} Agent modeling has explored a variety of modeling paradigms, including reinforcement learning approaches \cite{Osborne2021ASO,xu-etal-2021-generalization-text}, combined with
reading comprehension techniques \cite{narasimhan-etal-2015-language,tamari-etal-2019-playing,guo-etal-2020-interactive,yao-etal-2020-keep,yao-etal-2021-reading},
commonsense reasoning \cite{ryu-etal-2022-fire,murugesan-etal-2021-efficient},
graph-based networks \cite{ammanabrolu-riedl-2019-playing},
and neurosymbolic logic ~\cite{Kimura2021NeuroSymbolicRL,chaudhury-etal-2021-neuro,kimura-etal-2021-loa}.
Most recent agents make use of large pretrained language models \citep[e.g.][]{devlin-etal-2019-bert}, though these can pose challenges both in inference speed, as well as generalization to interactive environments.  
For example, a model that can correctly answer 90\% of multiple choice elementary science exam questions fails to solve text games that test that same knowledge but in a step-by-step procedural setting, even with significant training \cite{Wang2022ScienceWorldIY}.  

{\flushleft\textbf{Simulation Speed:}} A variety of simulators currently exist for text games, typically focusing on providing domain-general tooling for creating small procedurally generated research environments \cite[e.g.][]{Ct2018TextWorldAL}, or interfacing to the existing body of large interactive fiction games such as Zork \cite{lebling1979zork} from the 1980s and 1990s by providing tooling and APIs \cite{Hausknecht2020InteractiveFG}. Nearly all frameworks ultimately generate and run games as \textit{Z-machine} code \citep[e.g.][]{nelson2014zmachine}, an almost 40-year-old domain specific language designed for portability rather than simulation speed.  One of the central challenges in building fast research tooling is \textit{valid action generation}.  Because games implement different sets of actions, and at different levels of granularity, nearly all contemporary agents require the simulator to supply a list of possible valid actions (such as \textit{put coat in closet}) that could be undertaken by the agent at a given time step.  Action spaces can be large -- hundreds of thousands of action-object combinations are frequently possible at a given step in most games -- and existing frameworks \citep[e.g.][]{Hausknecht2020InteractiveFG} built on legacy tooling perform valid action generation by enumerating then running all possible action-object combinations at each game step then recording which ones are valid.  This is extremely costly, substantially reducing simulation performance (as shown in Table~\ref{tab:speed-comparison}).  In this work, \textsc{TextWorldExpress} has been built from the ground-up using heavily optimized and profiled code to quickly render environments while simultaneously generating an exhaustive list of possible next valid actions for agents, greatly speeding simulation time.  

\begin{table}[t!]
\begin{center}
\scriptsize
\setlength{\tabcolsep}{3pt}
\begin{tabular}{p{0.98\linewidth}} 
\toprule
    \textbf{CookingWorld Game} \\
\midrule
\textbf{North of House} \\
You are in the kitchen. In one part of the room you see a fridge that is closed. There is also a dining chair, that has nothing on it. You also see a dishwasher that is closed. In another part of the room you see a trash can that is closed. In one part of the room you see an oven. There is also a cutlery drawer that is closed. You also see a stove. In another part of the room you see a counter that has a \textbf{cookbook}, a \underline{green apple}, and a \underline{red hot pepper} on it. In one part of the room you see a kitchen cupboard that is closed. 
To the South you see the backyard. To the West you see the pantry. \\
\textit{> read \textbf{cookbook}} \\ 
~\\
Gather all following ingredients and follow the directions to prepare this tasty meal.

Ingredients:
  \textbf{purple potato}, 
  \textbf{yellow bell pepper}.

Directions:
  dice the \textbf{purple potato},
  fry the \textbf{purple potato},
  roast the \textbf{yellow bell pepper},
  prepare meal.

\textit{> open cutlery drawer} \\
~\\
You open the cutlery drawer. The cutlery drawer contains a \textbf{knife}. \\
\textit{> take \textbf{knife}} \\
~\\
You take the knife. \\
\textit{> move south} \\
~\\
You are in the backyard. In one part of the room you see a barbeque. There is also a patio chair, that has nothing on it. You also see a clothes line, that has nothing on it. In another part of the room you see a garden that has a raw \textbf{purple potato}, and a \textbf{yellow bell pepper} on it. In one part of the room you see a workbench, that has nothing on it. There is also a patio table, that has nothing on it. 
To the North you see the kitchen. \\
\textit{> take \textbf{purple potato}}\\
~\\

You take the \textbf{purple potato}.\\
\textit{> take \textbf{yellow bell pepper}}\\
~\\
You take the \textbf{yellow bell pepper}.\\
\textit{> move north} \\
~\\
You are in the kitchen.\\
\textit{> dice \textbf{purple potato}} \\
~\\
You dice the \textbf{purple potato}.\\
\textit{> cook \textbf{purple potato} in stove} \\
~\\
You fry the \textbf{purple potato} with the stove.\\
\textit{> cook \textbf{yellow bell pepper} with the oven} \\
~\\
You roast the \textbf{yellow bell pepper} with the oven. \\
\textit{> prepare meal} \\
~\\
The meal has been added to your inventory.\\
\textit{> eat meal} \\
~\\
\textit{Game completed.}\\

\bottomrule

\end{tabular}
\caption{\footnotesize An example CookingWorld text game, requiring the agent to read a recipe, collect ingredients, and follow preparation instructions.  User or agent-entered \textit{actions} for the 12-step solution are italicized.  Relevant \textbf{task objects} are bolded, while \underline{distractor objects} are underlined.  Here, an easy-level parametric game was generated consisting of 3 locations \textit{(kitchen, backyard, pantry)}, 2 required ingredients \textit{(potato, bell pepper)}, and 2 distractor ingredients \textit{(apple, hot pepper)}. 
\label{tab:CookingWorld-example}}
\end{center}
\vspace{-6mm}
\end{table}

\section{Environments}

%We implement high-speed versions of \todo{three popular benchmark environments} for text-game research: 
\textsc{TextWorldExpress} offers high-speed versions of three popular benchmark environments frequently used in text game research, as well as a number of newer environments for evaluating specific reasoning competencies:

{\flushleft\textbf{CookingWorld:}} The CookingWorld environment \cite{Ct2018TextWorldAL} tasks agents with preparing meals by following the instructions in a recipe that is provided in the environment.\footnote{This task was used as part of the First TextWorld Problems competition (\url{https://aka.ms/ftwp}) and named by \cite{madotto-go-explore2020}.}  Agents must first collect required food ingredients (e.g. milk, bell pepper, flour, salt) that can be found in the environment in canonical locations (e.g. kitchen, pantry, supermarket, garden) and containers (e.g. fridge, cupboard).  Randomly generated recipes require agents to first use a knife to prepare food by \textit{slicing, dicing,} or \textit{chopping} a subset of ingredients, then additionally using an appropriate heating appliance to \textit{fry, roast,} or \textit{barbeque} the ingredients.  If all ingredients are prepared according to the recipe, the agent can use an action to \textit{prepare the meal}, and finally \textit{eat the meal} to complete the task successfully.  Task complexity can be controlled by varying the number of locations in the environment, the number of ingredients required for the recipe, and the number of distractor ingredients randomly placed in the environment that are not required for the recipe.  The recipes and environments are parametrically generated, with subsets of ingredients and specific preparations held out between training, development, and test sets to prevent overfitting.  An example CookingWorld task is shown in Table~\ref{tab:CookingWorld-example}.

{\flushleft\textbf{TextWorld Commonsense (TWC):}} Text game agents frequently learn the dynamics of environment -- such as the need to open a door before one can move through it -- from interacting with the environment itself, rather than using a pre-existing knowledge base of common sense facts or object affordances that would speed task learning.  TextWorld Commonsense \cite{Murugesan2021TextbasedRA} aims to evaluate agents on common sense knowledge that can not be directly learned from the environment by providing agents a clean-up task where the agent must place common household objects (e.g. \textit{a dirty dish}) in their canonical locations (e.g. \textit{the dishwasher}) that can be found in knowledge bases such as ConceptNet \cite{liu2004conceptnet,speer2017conceptnet}.  Separate lists of objects are used in the training, development, and test sets, meaning the agent can not learn object locations from the training set alone, and must rely on an external common sense knowledge base to perform well on the development and test sets.  \citet{Murugesan2021TextbasedRA} specify three task difficulty levels, with the easiest including a single location and object to put away, while the hard setting includes two location and up to 7 objects.  %The \textsc{TextWorldExpress} reimplementation allows specifying up to 11 rooms, and any number of task-relevant and distractor objects. 

{\flushleft\textbf{Coin Collector:}} Agents frequently find tasks such as object search, environment navigation, or pick-and-place tasks challenging \cite{shridhar2020alfworld}.  The Coin Collector game \cite{Yuan2018CountingTE} distills these into a single benchmark where an agent must explore a series of rooms to locate and pick up a single coin.  In the original implementation, the game map typically takes the form of a connected loop or chain, such that continually moving to new locations means the agent will eventually discover the coin -- while including medium and hard modes that add in one or more ``dead-end'' paths.  To control for environment difficulty across games, the \textsc{TextWorldExpress} reimplementation uses the same map generator across environments, and generates arbitrary home environments rather than connected loops.  The user maintains control of other measures of difficulty, including the total number of rooms, and the number of distractor objects placed in the environment. 

{\flushleft\textbf{Adding new games:}} 
New games can be added to \textsc{TextWorldExpress}, and 4 additional games that benchmark arithmetic, navigation, and neurosymbolic reasoning have been added since its initial release\footnote{See full list at \url{https://github.com/cognitiveailab/TextWorldExpress\#environments}.}.  Adding new games takes about a day of coding, which can be more effortful than using the domain-specific implementation languages of existing game engines \citep[e.g.][]{Ct2018TextWorldAL}.

%
% Action Space
%
\begin{table}[]
    \centering
    \footnotesize
    \begin{tabular}{ll}
    \toprule
         Action & Description \\
         \midrule
            \multicolumn{2}{l}{\textit{Generic actions}}   \\
            \midrule
            look around                         &  \textit{describe current location}  \\
            inventory                           &  \textit{list agent inventory}  \\
            examine \textbf{OBJ}                &  \textit{examine an object}  \\
            move \textbf{DIR}                   &  \textit{move north, east, south, or west} \\
            open \textbf{OBJ}                   &  \textit{open a door or container} \\
            close \textbf{OBJ}                  &  \textit{close a door or container} \\
            take \textbf{OBJ}                   &  \textit{pick up an object} \\
            put \textbf{OBJ} in \textbf{OBJ}    &  \textit{put an object in a container} \\
            \midrule
            \multicolumn{2}{l}{\textit{Extended actions (CookingWorld)}}   \\
            \midrule
            read \textbf{OBJ}                   &  \textit{read a recipe book} \\
            cook \textbf{OBJ} in \textbf{OBJ}   &  \textit{cook an ingredient} \\
            chop \textbf{OBJ}                   &  \textit{chop an ingredient} \\
            slice \textbf{OBJ}                  &  \textit{slice an ingredient} \\    
            dice \textbf{OBJ}                   &  \textit{dice an ingredient}  \\    
            eat \textbf{OBJ}                    &  \textit{eat an ingredient} \\
            prepare meal                        &  \textit{prepare the meal} \\
            
         \bottomrule
    \end{tabular}
    \caption{The action space of the environments, as well as descriptions of each action.  Actions can take zero, one, or two object (OBJ) or direction (DIR) arguments. }
    \label{tab:action-space}
    \vspace{-4mm}
\end{table}

\subsection{Action Space and Valid Action Generation}

The three benchmark games each have up to 15 different types of actions available to agents, described in Table~\ref{tab:action-space}.  These include common text-game actions such as \textit{taking objects}, \textit{moving locations}, and \textit{opening doors}, as well as domain-specific actions such as \textit{slicing} or \textit{cooking ingredients} for the cooking-domain game.  Actions may take zero (e.g. \textit{look around}), one (e.g. \textit{take shirt}), or two (e.g. \textit{put shirt in closet}) objects as arguments.  

Most contemporary high-performance game agents \citep[e.g.][]{ammanabrolu2020graph,Murugesan2021TextbasedRA} make use of a ``valid-action handicap'' -- that is, at each step, they require a list of possible valid actions that can be taken in the environment, from which they select a single action to undertake.  For example, a kitchen agent might wish to \textit{dice the carrot}, but such an action would only be available to the agent if it currently possessed both a carrot and a knife in its inventory.  This valid-action detection is typically implemented overtop of existing interactive fiction games (such as \textit{Zork}) by an interface framework \citep[e.g., Jericho;][]{Hausknecht2020InteractiveFG} at significant loss to the simulation framerate.  In contrast, \textsc{TextWorldExpress} was designed from the ground-up to provide fast valid action generation to maintain high framerates.% \todo{PJ note, this paragraph is slightly repetative to some info in the related work -- we might be able to trim a bit.}

%\todo{TODO: Average number of valid actions per step, across the three games, for a random agent?}

%
%  Map Generation
%
\begin{figure}[!t]
%	\vspace{-12pt}
	\centering
	\includegraphics[scale=0.65]{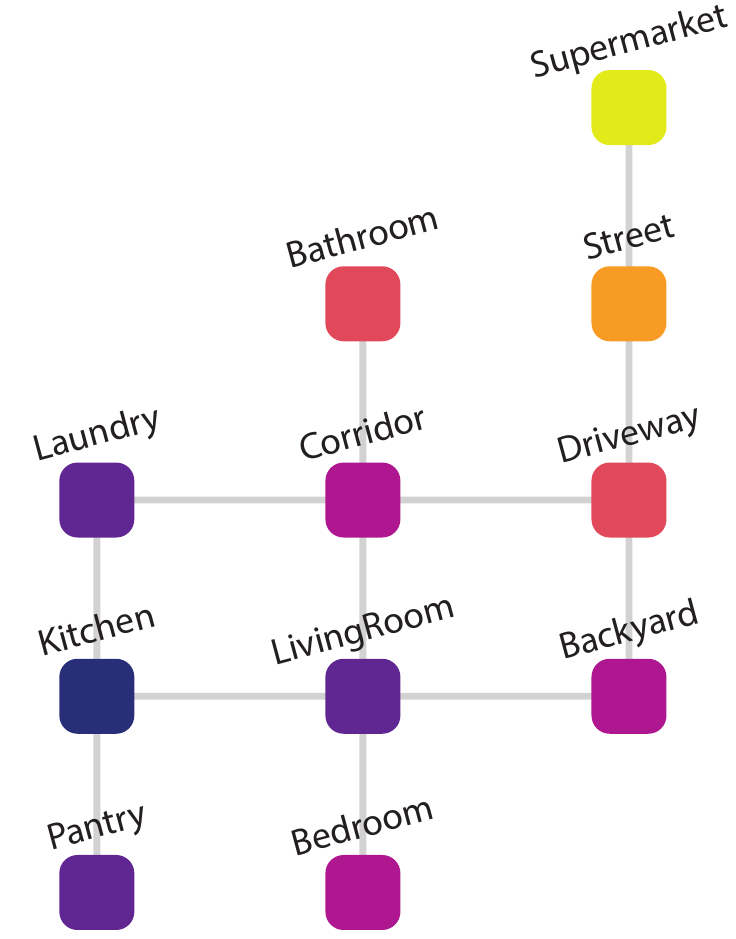}
%	\vspace{-20pt}
	\caption{An example of the random map generation process, containing 11 separate locations.  Locations are iteratively placed on a 7x7 grid, then interconnected (edges) on the four cardinal directions \textit{(north, east, south, west)} based on connection preferences.  For example, the \textit{Pantry} prefers to connect to the \textit{Kitchen}, and will never connect to the \textit{Bedroom}. }
	\label{fig:map-example}
	%\vspace{-14pt}
	\vspace{-4mm}
\end{figure}

\subsection{Map Generation}
Navigation tasks -- such as exploring an environment, or navigating to a specific location -- are typically challenging for contemporary text game agents.  Because of this,  games typically reduce the burden of navigation by providing simplified maps.  At one end of the extreme, the original \textit{TextWorld Commonsense} uses small maps containing only one or two locations, while at the other extreme \textit{CookingWorld} creates maps with over a dozen locations interconnected in common ways (i.e. a \textit{kitchen} is usually connected to a \textit{pantry}, \textit{backyard}, and/or \textit{corridor}, but is never directly connected to a \textit{supermarket}).  
%\textit{Coin Collector} allows generating maps that are rings, meaning that an agent could trivially solve the navigation task by moving to the next unseen location.  
To control for the difficulty of the navigation task across environments, \textsc{TextWorldExpress} uses the same map generator across all three benchmark games, while allowing the user to specify parameters such as the number of map locations to control the difficulty of the navigation task. 

Environments can consist of up to 11 locations, consisting of locations common to both the \textit{TWC} and \textit{CookingWorld} games.  Maps are randomly generated at the start of each game, and allow navigation on four cardinal directions (north, south, east, west).  Optionally, rooms may be connected with doors that an agent is required to open before allowing passage, increasing task complexity.  Figure~\ref{fig:map-example} shows an example map produced by the generator.

%- Include map generation figure
%- Same map generator used across environments, to control for the difficulty of the navigation task across environments (e.g. as the original Coin Collector used maps that were just loops). 

\subsection{Object Library}
Task objects, room objects, and distractor objects are populated from the object libraries provided by the TextWorld Commonsense and CookingWorld games.  This results in approximately 500 possible objects that can populate environments, including containers (e.g. \textit{fridge}, \textit{shelf}, \textit{countertop}), and movable objects (e.g. \textit{red onion, dirty shirt}). 
%- Objects from TWC
%- Also objects from CookingWorld

\subsection{Parametric Variation}

To reduce overfitting, generated tasks and environments vary in their requirements and presentation.  Tasks typically vary in task-critical objects, such as the specific objects that need to be cleaned up in \textit{TextWorld Commonsense}, or the recipe, ingredients, and their locations in \textit{CookingWorld}.  Environments parametrically vary, centrally in the environment map (how the rooms are interconnected), while also allowing different numbers of distractor objects to be generated in different randomized locations in the environment.  Critically, games are deterministic and the generation is repeatable and controlled by a single random seed, such that the same game can be regenerated during agent training and evaluation.  To create independent train, development, and test sets, in addition to each game having specific task objects that are unique across training and evaluation sets, we also assign blocks of random seeds to the train, development, and sets.  This allows generating thousands of possible parametric variations for each set, while ensuring that the tasks and environments remain unique.  

\subsection{Scoring}
At each time step, the simulator provides the agent a score that signifies the agent's progress in solving a given task.  Games typically assign rewards for critical task steps, such as picking up correct ingredients, or preparing ingredients correctly.  Because the total score required to complete a game can vary both across games and across task complexity, scores are provided both as raw counts, as well as normalized to between zero (no task progress) and one (task completion).  Each game has specific success and failure criterion, which are automatically detected by the simulator, and provided to the agent by the API.  For example, if a recipe requires a \textit{carrot} to be \textit{chopped}, but the agent instead \textit{slices} it, this will cause a task failure, and can be used as a reward signal for the agent model to use in adjusting its action policy.

\section{Speed Comparison}

\subsection{Online and Precrawled Generation}
%{\flushleft\textbf{Online and Precrawled generation:}} 
To enable extremely fast simulations, \textsc{TextWorldExpress} supports two game generation modes: normal (online) generation, and precrawled generation.  In \textit{online generation}, games are parametrically generated and played at runtime, allowing a large number of parametric game variations to be generated, and games to be played up to any number of steps.  Conversely, where speed is of critical importance, the simulator supports \textit{precrawling} all possible paths an agent might take in a given environment, and pre-caching these to disk as a JSON file.  This allows extremely fast game playing -- at essentially the speed of updating a pointer to a particular node in the precrawled state tree -- at the expense of generating and loading large files, that pragmatically limit the total number of steps that can be crawled and precached in the environment.\footnote{As an example, a 1GB file can typically store precrawled game trees for a single game variation up to between 8 and 12 steps, depending on the complexity of the action space.}  Precrawling is a unique feature offered by \textsc{TextWorldExpress}, as games taking minutes to crawl in this framework can take days or weeks to crawl in \textsc{TextWorld}.  

\subsection{Evaluating Simulation Speed}
%{\flushleft\textbf{Evaluating Simulation Speed}:}} 
We empirically compare the simulation speed of \textsc{TextWorldExpress} with three frameworks.

{\flushleft\textbf{TextWorld} \cite{Ct2018TextWorldAL}} is a framework for generating parametric text games for natural language processing research.  Games are specified using predicate logic (to define action rules) and a context-free grammar (to generate text), which TextWorld reformulates into Inform7 code \cite{nelson2006natural}, that is then ultimately compiled to a Z-Machine game \cite{nelson2014zmachine}.  The three benchmark games reimplemented in \textsc{TextWorldExpress} were originally implemented in TextWorld. 

{\flushleft\textbf{Jericho} \cite{Hausknecht2020InteractiveFG}} provides a research interface to the existing body of interactive fiction games, such as Zork \cite{lebling1979zork}, that were originally written for the Z-Machine interpreter.  Critically, Jericho provides facilities for action template extraction and valid-action generation, to reduce the difficulty of interfacing classic interactive fiction games with language agents. 

{\flushleft\textbf{ScienceWorld} \cite{Wang2022ScienceWorldIY}} is a science-domain text game simulator that provides the ability to train and evaluate agents on scientific tasks normally learned by elementary science students, such as changes of states of matter (melting, boiling, freezing), life cycles of plants and animals, and basic chemistry.  Supporting this is a series of complex simulation engines (e.g., thermodynamics, electrical conductivity, genetics) which increase simulation fidelity at the cost of speed.  Similar to TextWorld and Jericho, ScienceWorld supports generating valid actions at each time step. 

The results of this evaluation, using random agents to traverse the environments, are shown in Table~\ref{tab:speed-comparison}.  The highly optimized \textsc{TextWorldExpress} is able to simulate games in \textit{online} generation mode at an average of \textit{212k} frames per second per thread, or nearly three orders of magnitude faster than other frameworks.\footnote{\textsc{Python} performance is 10X slower than \textsc{Java/Scala} performance due to the speed of \textsc{Python-JVM} binders.} %  This is still two orders of magnitude faster than current simulators.}  
This varies between \textit{256k} steps per second for the fastest environment with the least complex action space \textit{(Coin Collector)}, to \textit{155k} steps per second for the most complex action space \textit{(CookingWorld)}.  On an 8-core workstation, this enables million-step experiments to be simulated per second, with billion-step experiments possible in approximately one hour.\footnote{%In practice in preliminary light-weight machine learning experiments using pre-crawled paths exported from \textsc{TextWorldExpress}, 
Using pre-crawled paths, we managed to run \textit{billion-game} experiment on a 32-core server in about a day.}  In contrast, one billion steps would take approximately 38 days using the original TextWorld implementations.  In \textit{precrawled} mode, where game states are precached, single-thread speeds of up to 4 million steps per second are possible.  Our fastest multi-threaded benchmark on desktop hardware (an \textsc{AMD 3950X} 16-core, 32-thread CPU) reaches 34M steps per second, enabling billion-step-scale simulations in approximately 30 seconds.

%\todo{-discuss Scala vs Python speed differences?}
%\todo{-discuss fastest mode (precrawled in a fast language like Java/scala -- ~4M frames/sec.}

%\todo{- describe measurement parameters: 
%- Jericho: All Jericho Suite (including Zork, etc).
%- TextWorld: NeverEnding (300 sps), Kitchen (150 sps). 
%- ScienceWorld: TODO
%- TextWorldExpress: Each single game using default parameters, as described above. 
%}

%\todo{- Mention that precrawling is not possible with existing environments beyond 4-5 steps, because they are too slow, where as here we bring it into the 12+ step range within minutes, which is within the solution range for many of these tasks). }

%\section{Modeling}
%- Genetic Algorithm example  (remove for space?)
%\todo{\lipsum[1]}

\section{Conclusion}
We present \textsc{TextWorldExpress}, a fast simulator for text-game research that reimplements three benchmark environments while running three orders of magnitude faster than their original implementations.  New games can be added using existing games as templates, and four new games benchmarking specific reasoning competencies like arithmetic and navigation have been added since its initial release.  The simulator supports common features (such as valid action detection), while providing new enabling features, such as quickly precrawling entire game state trees.  This work is released as open source. 

\section{Broader Impacts}
Embodied agents require a variety of common-sense reasoning skills and competencies about the world in order to successfully perform tasks.  Text games distill task learning to a high level of abstraction, allowing conceptual-level procedural knowledge to be acquired without simultaneously learning challenging low-level perceptual or motor tasks as in 3D simulators \citep[e.g.][]{ALFRED20,petrenko2021megaverse}, while reducing the computational requirements to run experiments from expensive GPU servers to common desktop hardware.  Futher, Shirdhar et al. \shortcite{shridhar2020alfworld} have empirically demonstrated that agents can be inexpensively pretrained on tasks in a text world environment, then transfer much of their performance to more realistic 3D environments, speeding  training.  \textsc{TextWorldExpress}, which increases the speed of text game experiments by three orders of magnitude, enables running experiments faster, at greater scale, or using alternate sample-heavy machine learning frameworks than currently available simulators.

\section{Limitations}
\textsc{TextWorldExpress} has two main limitations compared to existing simulators.  
\textsc{TextWorldExpress} gains much of its speed by developing a highly-profiled simulator with hard-coded implementations of text games.  Unlike the original \textsc{TextWorld} simulator, which is designed to allow new environments to be implemented with a domain-specific language, adding new environments to \textsc{TextWorldExpress} is more effortful and requires coding in \textsc{Scala}, a derivative of \textsc{Java}.
Similarly, for speed, the \textsc{TextWorldExpress} user input parser is simplified, and it only recognizes valid actions as it presents them to the agent, without facilities for alternate surface forms, misspellings, or other variations.  While it is common for agents to select actions from a valid action list, the lack of a diverse input parser limits utility for human participants who might choose to play these games.

\section*{Acknowledgements}
This work supported in part by National Science Foundation (NSF) award \#1815948 to PJ, and gift from the Allen Institute for Artificial Intelligence (AI2).

\bibliography{anthology,custom}
\bibliographystyle{acl_natbib}

% PJ NOTE: I don't think an Appendix is allowed in System Demo papers
%\appendix
%\section{Example Appendix}
%\label{sec:appendix}
%This is a section in the appendix.

\end{document}